\documentclass[10pt,twocolumn,letterpaper]{article}

\usepackage{cvpr}              

\newcommand{\Hypernerf}{HyperNeRF~\cite{park2021hypernerf} }
\newcommand{\Neud}{Neu3D~\cite{li2022neural} }
\newcommand{\Langsplat}{LangSplat~\cite{qin2024langsplat} }
\newcommand{\Featuredgs}{Feature-3DGS~\cite{Zhou_2024_CVPR} }
\newcommand{\Gaussiangrouping}{Gaussian Grouping~\cite{ye2025gaussian}}

\newcommand{\draftonly}[1]{#1}
\renewcommand{\draftonly}[1]{}

\newcommand{\jz}[1]{\draftonly{{\color{blue}[JZ: #1]}}}

\definecolor{cvprblue}{rgb}{0.21,0.49,0.74}
\usepackage[pagebackref,breaklinks,colorlinks,allcolors=cvprblue]{hyperref}
\usepackage{multirow}
\usepackage{amsmath}
\usepackage{amssymb}
\usepackage[accsupp]{axessibility}  

\usepackage{algorithm}
\usepackage{algpseudocodex}
\usepackage{booktabs}
\usepackage{makecell}
\usepackage{multirow}
\usepackage{booktabs}
\usepackage{bbding}
\usepackage{makecell}
\usepackage{adjustbox}
\def\paperID{6600} 
\def\confName{CVPR}
\def\confYear{2025}

\usepackage{bm}

\title{4D LangSplat: 4D Language Gaussian Splatting via Multimodal Large Language Models}

\author{
Wanhua Li$^{1,\ast}$, \quad  Renping Zhou$^{1, 2,\ast}$,  \quad  Jiawei Zhou$^{3}$, \quad Yingwei Song$^{1,4}$,    \quad  Johannes Herter$^{1,5}$, \\ \quad   Minghan Qin$^{2}$, \quad 
 Gao Huang$^{2,}$\textsuperscript{\Envelope}, \quad  Hanspeter Pfister$^{1,}$\textsuperscript{\Envelope}\\
  $^1$Harvard  University 
  $^2$Tsinghua University 
  $^3$Stony Brook University 
  $^4$Brown University 
  $^5$ETH Z\"urich
  \vspace{.2cm} \\
  \texttt{Project page: \url{https://4d-langsplat.github.io/}}
}

\begin{document}

\twocolumn[{%
\renewcommand\twocolumn[1][]{#1}%
\maketitle
\begin{center}
    \centering
    \captionsetup{type=figure}
    \includegraphics[width=1\textwidth]{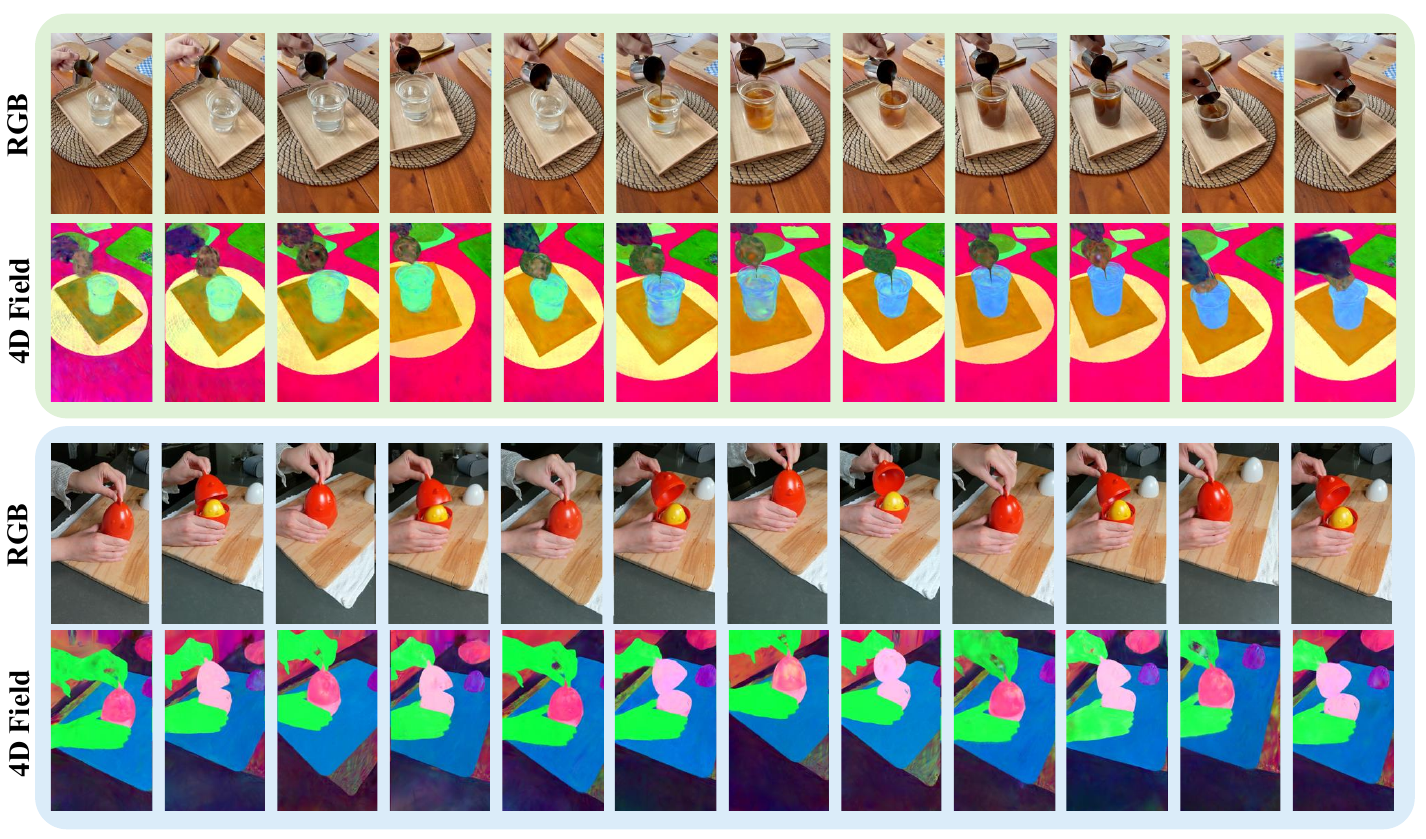}
    \vspace{-22pt}
    \captionof{figure}{Visualization of the learned language features of our 4D LangSplat.
    We observe that 4D LangSplat effectively learns dynamic semantic features that change over time, such as the gradual diffusion of coffee shown in the first two rows, and the ``chicken'' toggling between open and closed states in the latter two rows. Additionally, our semantic field captures consistent features for semantics that remain unchanged over time, with the clear object boundaries in the visualization demonstrating the precision of our semantic field.}
    \vspace{-3pt}
    \label{fig:teaser}
\end{center}%
}]
{\let\thefootnote\relax\footnotetext{{$^{\ast}$ Equal contribution.\ \textsuperscript{\Envelope}Corresponding authors.}}}

\maketitle
\begin{abstract}
Learning 4D language fields to enable time-sensitive, open-ended language queries in dynamic scenes is essential for many real-world applications. While LangSplat successfully grounds CLIP features into 3D Gaussian representations, achieving precision and efficiency in 3D static scenes, it lacks the ability to handle dynamic 4D fields as CLIP, designed for static image-text tasks, cannot capture temporal dynamics in videos. Real-world environments are inherently dynamic, with object semantics evolving over time. Building a precise 4D language field necessitates obtaining pixel-aligned, object-wise video features, which current vision models struggle to achieve. To address these challenges, we propose 4D LangSplat, which learns 4D language fields to handle time-agnostic or time-sensitive open-vocabulary queries in dynamic scenes efficiently. 4D LangSplat bypasses learning the language field from vision features and instead learns directly from text generated from object-wise video captions via Multimodal Large Language Models (MLLMs). Specifically, we propose a multimodal object-wise video prompting method, consisting of visual and text prompts that guide MLLMs to generate detailed, temporally consistent, high-quality captions for objects throughout a video. These captions are encoded using a Large Language Model into high-quality sentence embeddings, which then serve as pixel-aligned, object-specific feature supervision, facilitating open-vocabulary text queries through shared embedding spaces. Recognizing that objects in 4D scenes exhibit smooth transitions across states, we further propose a status deformable network to model these continuous changes over time effectively. Our results across multiple benchmarks demonstrate that 4D LangSplat attains precise and efficient results for both time-sensitive and time-agnostic open-vocabulary queries.
\end{abstract}    
\section{Introduction}
\label{sec:intro}

The ability to construct a language field~\cite{kerr2023lerf,qin2024langsplat} that supports open vocabulary queries holds significant promise for various applications such as robotic navigation~\cite{huang2023visual}, 3D scene editing~\cite{kobayashi2022decomposing}, and interactive virtual environments~\cite{liu2023weakly}. 
Due to the scarcity of large-scale 3D datasets with rich language annotations, current methods~\cite{kerr2023lerf,shen2023distilled,liu2023weakly} leverage pre-trained models like CLIP~\cite{radford2021learning} to extract pixel-wise features, which are then mapped to 3D spaces.
Among them, LangSplat~\cite{qin2024langsplat} received increasing attention due to its efficiency and accuracy, which grounds the precise masks generated by the Segment Anything Model (SAM)~\cite{kirillov2023segment} with CLIP features into 3D Gaussians, achieving an accurate and efficient 3D language field by leveraging 3D Gaussian Splatting (3D-GS)~\cite{kerbl20233d}.  LangSplat supports open-vocabulary queries in various semantic levels by learning three SAM-defined semantic levels.

Nothing endures but change. 
Real-world 3D scenes are rarely static, and they continuously change and evolve. 
To enable open-vocabulary queries in dynamic 4D scenes, it is crucial to consider that target objects may be in motion or transformation. For instance, querying a scene for ``\emph{dog}" in a dynamic environment may involve the dog running, jumping, or interacting with other elements. Beyond spatial changes, users may also want time-related queries, such as ``\emph{running dog}", which should only respond during the time segments when the dog is indeed running. Therefore, supporting time-agnostic and time-sensitive queries within a 4D language field is essential for realistic applications.

A straightforward approach to extend LangSplat to a 4D scene is to learn a deformable Gaussian field~\cite{wu20244d,yang2024deformable,li2024spacetime} with CLIP features. However, it cannot model the dynamic, time-evolving semantics as CLIP, designed for static image-text matching~\cite{li2022ordinalclip,ding2024tree}, struggles to capture temporal information such as state changes, actions, and object conditions~\cite{tong2024eyes,tong2024cambrian}. 
Learning a precise 4D language field would require \emph{pixel-aligned, object-level} video features as the 2D supervision to capture the spatiotemporal semantics of each object in a scene, yet current vision models~\cite{wang2023internvid,xu2021videoclip} predominantly extract \emph{global, video-level} features. One could extract features by cropping interested objects and then obtain patch features. It inevitably includes background information, leading to imprecise semantic features~\cite{qin2024langsplat}. Removing the background and extracting vision features only from the foreground object with accurate object masks leads to ambiguity in distinguishing between object and camera motion, since only the precise foreground objects are visible without a reference to the background context. These pose significant challenges for building an accurate and efficient 4D language field.

To address these challenges, we propose 4D LangSplat, which constructs a precise and efficient 4D Language Gaussian field to support time-agnostic and time-sensitive  open-vocabulary queries. We first train a 4D Gaussian Splatting (4D-GS)~\cite{wu20244d} model to reconstruct the RGB scene, which is represented by a group of Gaussian points and a deformable decoder defining how the Gaussian point changes its location and shape over time. Our 4D LangSplat then enhances each Gaussian in 4D-GS with two language fields, where one learns time-invariant semantic fields with CLIP features as did in LangSplat, and the other learns time-varying semantic field to capture the dynamic semantics. The time-invariant semantic field encodes semantic information that does not change over time such as ``\emph{human}", ``\emph{cup}", and ``\emph{dog}". They are learned with CLIP features on three SAM-defined semantic levels. 

For the time-varying semantic field, instead of learning from vision features, we propose to directly learn from textual features to capture temporally dynamic semantics. Recent years have witnessed huge progress~\cite{openai2023gpt4v,team2024gemini} of Multimodal Large Language Models (MLLMs), which take multimodal input, including image, video, and text, and generate coherent responses. 
Encouraged by the success of MLLMs, we propose a multimodal object-wise video prompting method that combines visual and text prompts to guide MLLMs in generating detailed, temporally consistent, high-quality captions for each object throughout a video. We then encode these captions using a large language model (LLM) to extract sentence embeddings, creating pixel-aligned, object-level features that serve as supervision for the 4D Language field.
Recognizing the smooth transitions exhibited by objects across states in 4D scenes, we further introduce a status deformable network to model these continuous state changes effectively over time. Our network captures the gradual transitions across object states, enhancing the model’s temporal consistency and improving its handling of dynamic scenes. Figure \ref{fig:teaser} visualizes the learned time-varying semantic field. Our experiments across multiple benchmarks validate that 4D LangSplat achieves precise and efficient results, supporting both time-agnostic and time-sensitive open-vocabulary queries in dynamic, real-world environments.

In summary, our contributions are threefold: 
\begin{itemize}
\item We introduce 4D LangSplat for open-vocabulary 4D spatial-temporal queries. To the best of our knowledge, we are the first to construct 4D language fields with object textual captions generated by MLLMs.
\item To model the smooth transitions across states for objects in 4D scenes, we further propose a status deformable network to capture continuous temporal changes.
\item Experiential results show that our method attains state-of-the-art performance for both 
time-agnostic and time-sensitive open-vocabulary queries.
\end{itemize}

\section{Related Work}
\label{sec:related_work}

\textbf{3D Gaussian Splatting.}
3D-GS~\cite{kerbl20233d} is a powerful volumetric rendering technique that has gained attention for its real-time, high-quality rendering ability. It represents complex surfaces and scenes by projecting 3D Gaussian distributions into 2D image space. It has been widely used for many applications such as human reconstruction~\cite{li2024gaussianbody,shao2024splattingavatar}, 3D editing~\cite{chen2024gaussianeditor,wang2024gaussianeditor}, mesh extraction~\cite{waczynska2024games,gao2024mesh}, autonomous driving~\cite{zhou2024drivinggaussian,zhou2024hugs}. Recent work~\cite{yang2024deformable,li2024spacetime,lin2024gaussian,bae2024per} including 4D Gaussian Splatting (4D-GS)~\cite{wu20244d} has extended Gaussian Splatting to 4D by introducing deformable fields, allowing for dynamic scenes where Gaussian parameters evolve over time to capture both spatial and temporal transformations. However, 4D-GS primarily focuses on visual fidelity rather than semantic understanding, which limits its applicability in open-vocabulary language queries.

\noindent\textbf{3D Language Field.} Some early work~\cite{tschernezki2022neural,kobayashi2022decomposing} usually ground 2D foundation model features~\cite{radford2021learning,caron2021emerging,li2022language} into a neural radiance field (NeRF)~\cite{mildenhall2021nerf}. For example, Distilled Feature Fields (DFFs) propose to distill CLIP-LSeg~\cite{li2022language} into NeRF for semantic scene editing. LERF~\cite{kerr2023lerf} proposes to distill CLIP~\cite{radford2021learning} features into NeRF to support open-vocabulary 3D querying. With the emergence of 3D-GS, many methods~\cite{shi2024language,Zhou_2024_CVPR,ye2025gaussian,yu2024language} adopt 3D-GS as the 3D scene representation and lift 2D foundation model features into 3D Gaussians. Among them, LangSpalt~\cite{qin2024langsplat} attains precise and efficient language fields due to the introduction of SAM masks. By incorporating multiple levels of semantic granularity, LangSplat effectively supports open-vocabulary queries across whole objects, parts, and subparts. 
Recently, several methods have attempted to embed semantic fields in 4D scenes and have achieved promising progress, such as DGD~\cite{labe2024dgd} and 4-LEGS~\cite{fiebelman20244}. However, these approaches have not leveraged the powerful generative capabilities of Multimodal Large Language Models.


\noindent\textbf{Multimodal Large Language Models.} The remarkable success of LLMs~\cite{chiang2023vicuna,touvron2023llama,touvron2023llama2,bai2023qwen} has shown their ability to perform new tasks~\cite{li2024socialgpt} following human instructions. Based on LLMs, the research on MLLMs~\cite{openai2024gpt4o,bai2023qwenvl,liu2024improved} explores the possibility of multimodal chat ability~\cite{huang2024dynamic}, which represents a significant step forward in integrating visual and textual modalities for complex scene understanding. MLLMs usually employ a vision encoder to extract visual features and learn a connector to align visual features with LLMs. The recent models~\cite{wang2024qwen2,cheng2024videollama,li2024llava} demonstrate remarkable capabilities in generating coherent captions from multimodal inputs, including images and videos. In this paper, we propose to utilize the powerful multimodal process ability of MLLMs to convert video data into object-level captions, which are then used to train a 4D language field.

\section{Method}
\label{sec:method}

\subsection{Preliminaries}

\textbf{3D Gaussian Splatting.}  
In 3D-GS~\cite{kerbl20233d}, a scene is represented as a set of 3D Gaussian points.
Each pixel in 2D images is computed by blending $N$ sorted 3D Gaussian points that overlap the pixel:
\begin{equation}
    C = \sum_{i=1}^{N} c_i \alpha_i \prod_{j=1}^{i-1} (1 - \alpha_j),
    \label{eq:rendering_3dgs}
\end{equation}
where $c_i$ and $\alpha_i$ are the color and density of $i$-th Gaussian. 

\noindent\textbf{LangSplat.} Building upon 3D-GS, LangSplat~\cite{qin2024langsplat} grounds 2D CLIP features into 3D Gaussians. To obtain a precise field, SAM is used to obtain accurate object masks and then CLIP features are extracted with masked objects. 
LangSplat adopts feature splatting to train the 3D language field:
\begin{equation}
    \bm{F} = \sum_{i =1}^{N} \bm{f}_i \alpha_i \prod_{j=1}^{i-1} (1 - \alpha_j),
    \label{eq:rendering_langsplat}
\end{equation}
where $\bm{f}_i$ represents the language feature of the $i$-th Gaussians and $\bm{F}$ is the rendered embedding in 2D images.

\noindent\textbf{4D Gaussian Splatting.} 4D-GS~\cite{wu20244d} extends the 3D-GS for dynamic scenes by introducing a deformable Gaussian field. 
Here, Gaussian parameters, including position, rotation, and scaling factor, are allowed to vary over time:
\begin{equation}
    (\mathcal{X}', r', s')= (\mathcal{X} + \Delta\mathcal{X}, r + \Delta r, s + \Delta s),
    \label{eq:rendering_4dgs}
\end{equation}
where $\mathcal{X}$, $r$, and $s$ represent the position, rotation, and scaling parameters, respectively. $\Delta\mathcal{X}$, $\Delta r$, and $\Delta s$ denote the corresponding deformable networks, which are implemented by lightweight MLPs. The HexPlane~\cite{cao2023hexplane,fridovich2023k} representation is used to obtain rich 3D Gaussian features. 

A straightforward approach to adapting LangSplat for 4D scenes is to extend its static 3D language Gaussian field with a deformable Gaussian field, as done in 4D-GS. However, this approach faces significant limitations due to the nature of CLIP features. CLIP~\cite{radford2021learning} is designed primarily for static image-text alignment, making it ill-suited for capturing dynamic and time-evolving semantics in video. Recent research~\cite{wang2021actionclip,tong2024eyes,tong2024cambrian} further confirms that it struggles with understanding state changes, actions, object conditions, and temporal context. For a precise and accurate 4D language field, it is essential to obtain \emph{pixel-aligned, object-level} features that track temporal semantics with fine-grained detail for each object in a scene. However, existing vision models~\cite{wang2023internvid,xu2021videoclip} primarily offer \emph{global, video-level} features that overlook specific object-level information, making it difficult to represent spatiotemporal semantics at the object level. While cropping objects and obtaining patch-based features is possible, this includes background information, leading to inaccurate language fields. Further cropping objects with accurate masks makes it difficult for vision models to distinguish between object movement and camera motion, as there is no background reference.

\begin{figure*}[t]
    \centering
    \includegraphics[width=1\linewidth]{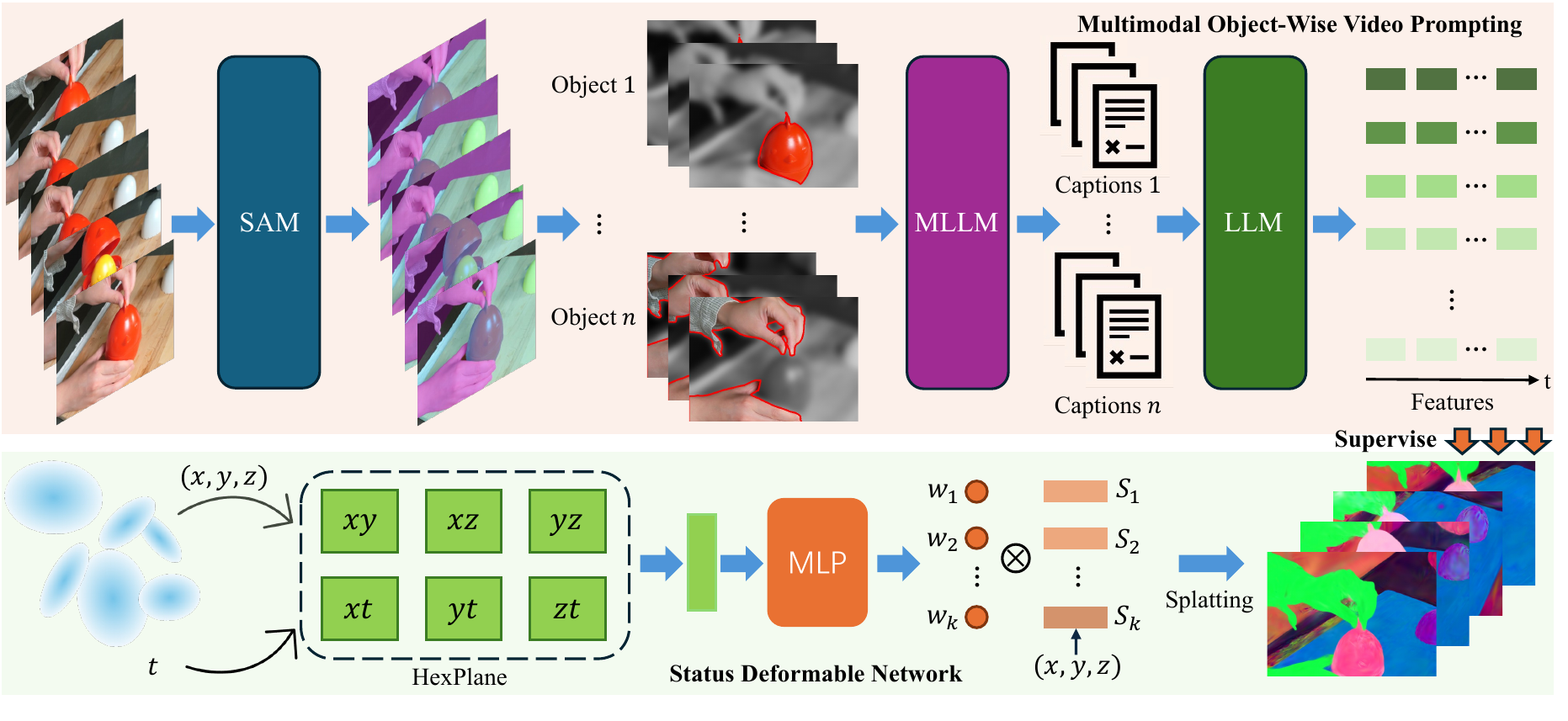}
    \caption{The framework of constructing a time-varying semantic field in 4D LangSplat. We first use multimodal object-wise prompting to convert a video into pixel-aligned object-level caption features. Then, we learn a 4D language field with a status deformable network.}
    \label{fig:framework}
\end{figure*}

\subsection{4D LangSplat Framework}

To address these challenges, we introduce 4D LangSplat, which constructs accurate and efficient 4D language fields to support both time-sensitive and time-agnostic open-vocabulary queries in dynamic scenes.
We first reconstruct the 4D dynamic RGB scene using 4D-GS~\cite{wu20244d}. In this stage, the RGB scene is represented by a set of deformable Gaussian points, each with parameters that adjust over time to capture object movement and shape transformations within the scene. Building on the learned 4D-GS model, we extend each Gaussian point with language embeddings to learn 4D language fields. To further capture temporal and spatial details, and to handle both time-sensitive and time-agnostic queries effectively, we simultaneously construct two types of semantic fields: a time-agnostic semantic field and a time-varying semantic field. The time-agnostic semantic field focuses on capturing semantic information that does not change over time. Although objects in the scene are dynamic, they still exhibit attributes that remain constant across time, such as static properties of entities like ``\emph{dog}", ``\emph{human}", and other objects within the environment. This semantic field emphasizes spatial details of these time-agnostic semantics. Conversely, the time-varying semantic field captures temporally dynamic semantics, such as ``\emph{a running dog}" 
, emphasizing semantic transitions over time.

For the time-agnostic semantic field, we still use CLIP features and lift them to 4D space, as they are sufficient for capturing time-agnostic semantics. Specifically, we learn a static language embedding for each deformable Gaussian point in the 4D-GS model. Similar to LangSplat, we utilize SAM’s hierarchical segmentation masks, learning three distinct time-agnostic semantic fields corresponding to the three levels of semantic granularity provided by SAM. Although each Gaussian point's position and shape dynamically change over time, its semantic feature remains static. These static embeddings ensure spatial accuracy while focusing on stable semantic information derived from CLIP features. 
On the other hand, to learn the time-varying semantic field, we propose a novel approach that bypasses the limitations of vision-based feature supervision. Instead, visual data is converted into object-level captions by leveraging MLLMs. These captions are then encoded using an LLM to extract sentence embeddings, which are used as pixel-aligned, object-level features for training the semantic field. To effectively model the smooth, continuous transitions of Gaussian points between a limited set of states, we further introduce a status deformable network to enhance reconstruction quality. The framework of training time-varying 4D fields is illustrated in Figure \ref{fig:framework}.

\subsection{Multimodal Object-Wise Video Prompting} 
\label{subsec:mllm-video-prompting}
Constructing a high-quality, dynamic 4D semantic field requires detailed, pixel-aligned object-level features that capture time-evolving semantics in video data. However, obtaining these fine-grained visual features is challenging due to the limitations of current vision models in distinguishing object-level details over time.  To overcome this, we propose converting video segments into object-wise captions and extracting sentence embeddings from these captions to serve as precise, temporally consistent features. 

Advances in MLLMs like GPT-4o~\cite{openai2024gpt4o}, LLaVA-OneVision~\cite{li2024llava}, and Qwen2-VL~\cite{wang2024qwen2} enable high-quality language generation from multimodal inputs. These models process video, image, and text inputs to generate temporally consistent responses. Leveraging these capabilities, we propose a multimodal object-wise video prompting method, which combines visual and textual prompts to guide the MLLM in generating temporally consistent, object-specific, high-quality captions across video frames, encapsulating both spatial and temporal details.

Formally, let \(V = \{I_1, I_2, \dots, I_T\}\) be a video segment of \(T\) frames. For each frame, we apply SAM~\cite{kirillov2023segment} in conjunction with DEVA tracking~\cite{cheng2023tracking} to segment objects and maintain consistent object identities over time. This process yields temporally consistent masks for \(n\) objects present in the video, denoted as \(\{M_1, M_2, \dots, M_n\}\), where each mask \(M_i\) represents a specific object tracked across frames. Each frame \(I_t\) is segmented with the object masks at time step $t$ \(\{M_{1,t}, M_{2,t}, \dots, M_{n,t}\}\).

To effectively generate instance-wise, object-specific captions while preserving the broader scene context, we need to guide the MLLM through precise prompting. Our goal is for the MLLM to generate captions focused solely on the target object without introducing details of other objects. However, the presence of other objects as background reference remains essential; without this context, the MLLM may lose track of spatial relationships and environmental context, which are critical for understanding the action and status of the target object. Thus, our approach employs prompting techniques to direct the MLLM’s attention to each object, enabling region-specific captioning that maintains overall scene awareness. Inspired by the recent visual prompting progress~\cite{subramanian2022reclip,shtedritski2023does,yang2024fine}, we first use visual prompts to highlight the object of interest. 
Specifically, we build a visual prompt \(\mathcal{P}_{i,t}\) for each object \(i\) in frame \(I_t\):
\begin{equation}
    \mathcal{P}_{i,t} = \operatorname{Contour}(M_{i,t}) \cup \operatorname{Gray}( M_{i,t}) \cup \operatorname{Blur}( M_{i,t}),
    \label{eq:visualprompt}
\end{equation}
where \(\operatorname{Contour}(M_{i,t})\) highlights \(M_{i,t}\) with a red contour, \(\operatorname{Gray}(M_{i,t})\) converts the non-object area to grayscale, and \(\operatorname{Blur}( M_{i,t})\) applies a Gaussian blur to the background pixels.  This prompt preserves essential background information while ensuring focus on the object of interest, improving the MLLM’s attention to the relevant target.

For temporal coherence, we first generate a high-level video-level motion description for object \(i\), noted as \(\mathcal{D}_i\), which summarizes the motion dynamics over \(T\) frames. This description is derived by prompting the MLLM with the entire video sequence \(V\) to capture object motion and interactions, defined as:
\begin{equation}
    \mathcal{D}_i = \operatorname{MLLM}(\{\mathcal{P}_{i,1},..., \mathcal{P}_{i,T}\},\mathcal{T}_{video}, V ),
    \label{eq:videocaption}
\end{equation}
where $\mathcal{T}_{video}$ denotes the textual prompt that instructs the MLLM to generate video-level motion descriptions based on the visual prompts.
This description \(\mathcal{D}_i\) is then used as context for generating frame-specific captions. For each frame \(I_t\), we combine \(\mathcal{D}_i\) with the visual prompt \(\mathcal{P}_{i,t}\) to generate a time-specific caption \(C_{i,t}\), capturing both the temporal and contextual details for object \(i\) in frame \(I_t\):
\begin{equation}
C_{i,t} = \operatorname{MLLM}(\mathcal{D}_i, \mathcal{P}_{i,t}, \mathcal{T}_{frame}, V_t ),
    \label{eq:framecaption}
\end{equation}
where $\mathcal{T}_{\text{frame}}$ denotes the textual prompt that instructs the MLLM to generate an object caption describing the object's current action and status at a specific time step.


Each caption \(C_{i,t}\) provides semantic information for an object $i$ at time $t$. To encode this semantic data into features for training the 4D language field, we extract sentence embeddings \(\bm{e}_{i,t}\) for each caption \(C_{i,t}\). As LLMs exhibit strong processing ability for free-form text~\cite{touvron2023llama,team2024gemma}, we further propose to utilize them to extract sentence embeddings. Specifically, a fined-tuned LLM~\cite{wang2023improving} for sentence embedding tasks is used to extract features. This design choice allows our model to respond effectively to open-vocabulary queries as the embeddings are generated within a shared language space that aligns with natural language queries. Thus, for every pixel \((x, y) \in M_{i,t}\) within object \(i\)'s mask in frame \(I_t\), the feature \(\bm{F}_{x,y,t}\) is given by:
\begin{equation}
\bm{F}_{x,y,t} = \bm{e}_{i,t},
    \label{eq:featurelabels}
\end{equation}
where the embeddings \(\bm{F}_{x,y,t}\) serve as 2D supervision for the time-variable semantic field, providing pixel-aligned, object-wise features across frames.

\subsection{Status Deformable Network}
\label{subsec:status-deformable}

With the 2D semantic feature supervision information available, we use it to train a 4D field. A straightforward approach, analogous to the method used 4D-GS, would be to directly learn a deformation field $\Delta \bm{f}$ for the semantic features of deformable Gaussian points. However, this straightforward approach allows the semantic features of each Gaussian point to change to any arbitrary semantic state, potentially increasing the learning complexity and compromising the temporal consistency of the features.
In real-world dynamic scenes, each Gaussian point typically exhibits a gradual transition between a limited set of semantic states. For instance, an object like a person may transition smoothly among a finite set of actions (\eg, standing, walking, running), rather than shifting to entirely unrelated semantic states. To model these smooth transitions and maintain a stable 4D semantic field, we propose a status deformable network that restricts the Gaussian point’s semantic features to evolve within a predefined set of states.

Specifically, we represent the semantic feature of a Gaussian point $i$ at any time \(t\) as a linear combination of \(K\) state prototype features, \(\{\bm{S}_{i,1}, \bm{S}_{i,2}, \dots, \bm{S}_{i,K}\}\), where each state captures a specific, distinct semantic meaning. The semantic feature $\bm{f}_{i, t}$ of a Gaussian point $i$ at time \(t\) is:
\begin{equation}
    \bm{f}_{i, t} = \sum_{k=1}^{K} w_{i,t,k} \bm{S}_{i,k},
    \label{eq:status_deformable_combination}
\end{equation}
where \(w_{i,t,k}\) denotes the weighting coefficient for each state \(k\) at time \(t\), with \(\sum_{k=1}^{K} w_{i,t,k} = 1\). This linear combination ensures that each Gaussian point’s semantic features transition gradually between predefined states.

To determine the appropriate weighting coefficients \(w_{k,t}\) for each Gaussian point over time, we employ an MLP decoder $\phi$. This MLP takes as input the spatial-temporal features from Hexplane~\cite{cao2023hexplane} and predicts weighting coefficients that reflect the temporal progression of semantic states. The MLP decoder $\phi$ and the per-Gaussian states \(\{\bm{S}_{i,1}, \bm{S}_{i,2}, \dots, \bm{S}_{i,K}\}\) are jointly trained. This design ensures that the status deformable network adapts to both the spatial and temporal context, enabling smooth, consistent transitions among semantic states.

\subsection{Open-vocabulary 4D Querying} 
After training, 4D LangSplat enables both time-agnostic and time-sensitive open-vocabulary queries. 
For time-agnostic queries, we utilize only the time-agnostic semantic field. We first render a feature image and then compute the relevance score~\cite{kerr2023lerf} between this rendered feature image and the query. Following the post-processing strategy in LangSplat~\cite{qin2024langsplat}, we obtain the segmentation mask for each frame from the relevance score maps. 

For time-sensitive queries, we combine both the time-agnostic and time-sensitive semantic fields. First, the time-agnostic semantic field is used to derive an initial mask for each frame, following the same procedure described above. This mask identifies where the queried object or entity exists, irrespective of time. To refine the query to specific time segments where the queried term is active (\eg, an action occurring within a particular timeframe), we calculate the cosine similarity between the time-sensitive semantic field on the initial mask region and the query text. This similarity is computed across each frame within the masked region to determine when the time-sensitive characteristics of the query term are most strongly represented. Using the mean cosine similarity value across the entire video as a threshold, we identify the frames that exceed this threshold, indicating relevant time segments. The spatial mask obtained with the time-agnostic field is retained as the final mask prediction for the identified time segments.

This combination of time-agnostic and time-sensitive semantic fields enables accurate and efficient spatiotemporal querying, allowing 4D LangSplat to capture both the persistent and dynamic characteristics of objects in the scene.

\section{Experiment}
\label{sec:experiments}



\begin{table*}[t]
\renewcommand\tabcolsep{6pt}
\centering
\resizebox{\textwidth}{!}{%
\begin{tabular}{lcccccccccc}
\toprule
\multirow{2}{*}{Method} & \multicolumn{2}{c}{americano} & \multicolumn{2}{c}{chickchicken} & \multicolumn{2}{c}{split-cookie} & \multicolumn{2}{c}{espresso} & \multicolumn{2}{c}{Average} \\
\cmidrule(lr){2-3} \cmidrule(lr){4-5} \cmidrule(lr){6-7} \cmidrule(lr){8-9} \cmidrule(lr){10-11}
 & Acc(\%) & vIoU(\%) & Acc(\%) & vIoU(\%) & Acc(\%) & vIoU(\%) & Acc(\%) & vIoU(\%) & Acc(\%) & vIoU(\%) \\
\midrule
\Langsplat & 45.19 & 23.16 & 53.26 & 18.20 & 73.58 & 33.08 & 44.03 & 16.15 & 54.01 & 22.65 \\
Deformable CLIP & 60.57 & 39.96 & 52.17 & 42.77 & 89.62 & 75.28 & 44.85 & 20.86 & 61.80 & 44.72 \\
Non-Status Field & 83.65 & 59.59 & 94.56 & 86.28 & 91.50 & 78.46 & 78.60 & 47.95 & 87.58 & 68.57 \\
\midrule
Ours & \textbf{89.42} & \textbf{66.07} & \textbf{96.73} & \textbf{90.62} & \textbf{95.28} & \textbf{83.14} & \textbf{81.89} & \textbf{49.20} & \textbf{90.83} & \textbf{72.26} \\
\bottomrule
\end{tabular}}
\vspace{-8pt}
\caption{Quantitative comparisons of time-sensitive querying on the \Hypernerf dataset.}
\vspace{-12pt}
\label{tab:dynamic-main-results}
\end{table*}

\begin{table}[t]
\renewcommand\tabcolsep{4pt}
  \centering
  \begin{tabular}{lcccc}
    \toprule
    \multirow{2}{*}{Method}  & \multicolumn{2}{c}{HyperNeRF} & \multicolumn{2}{c}{Neu3D} \\
    \cmidrule(lr){2-3} \cmidrule(lr){4-5} & mIoU & mAcc & mIoU & mAcc \\
    \midrule
    \Featuredgs & 36.63 & 74.02 & 34.96 & 87.12 \\
    \Gaussiangrouping & 50.49 & 80.92 & 49.93 & 95.05 \\
    \Langsplat & 74.92  & 97.72 &  61.49 & 91.89 \\
    \midrule
    Ours & \textbf{82.48} & \textbf{98.01} & \textbf{85.11} & \textbf{98.32} \\
    \bottomrule
  \end{tabular}
  \vspace{-6pt}
\caption{ Quantitative comparisons of time-agnostic querying on the \Hypernerf and \Neud datasets (Numbers in \%).}
  \vspace{-10pt}
  \label{tab:static-main-results}
\end{table}


\begin{table}[h!]
\renewcommand\tabcolsep{5pt}
\centering
\begin{minipage}{.2\textwidth}
    \centering
    \begin{tabular}{cccc}
    \toprule
    Blur & Gray & Contour & $\Delta_{\mathrm{sim}}$ \\
    \midrule
    $\checkmark$ & & & 0.33 \\
    $\checkmark$ & $\checkmark$ & & 2.15 \\
    $\checkmark$ & $\checkmark$ & $\checkmark$ & 3.32 \\
    \bottomrule
    \end{tabular}
    \vspace{-8pt}
    \caption{Comparisons of Visual prompts.}
    \vspace{-10pt}
    \label{tab:aba-visual-prompt}
\end{minipage}%
\hspace{35pt}
\begin{minipage}{.2\textwidth}
    \centering
    \begin{tabular}{ccc}
    \toprule
    Video & Image & $\Delta_{\mathrm{sim}}$ \\
    \midrule
    & & 0.14 \\
    $\checkmark$ & & 1.01 \\
    $\checkmark$ & $\checkmark$ & 3.32 \\
    \bottomrule
    \end{tabular}
    \vspace{-8pt}
    \caption{Comparisons of Text prompts.}
    \vspace{-10pt}
    \label{tab:aba-text-prompt}
\end{minipage}
\end{table}

\begin{table}[h!]
\renewcommand\tabcolsep{7pt}
\centering
\begin{tabular}{lccccc}
\toprule
K & 2 & 3 & 4 & 5 & 6 \\
\midrule
Acc (\%) & 94.56 & \textbf{97.82} & 95.65 & 94.56 & 94.56 \\
vIoU (\%) & 88.05 & \textbf{91.93} & 89.11 & 88.98 & 86.28 \\
\bottomrule
\end{tabular}
\vspace{-6pt}
\caption{Results for different state numbers on \textit{chick\textunderscore chicken}.}
\vspace{-20pt}
\label{table:ablation-centers}
\end{table}
\subsection{Setup}
\textbf{Datasets.} We conduct evaluations using two widely adopted datasets: \Hypernerf and \Neud. Given the absence of semantic segmentation annotations for dynamic scenes in these datasets, we perform manual annotations to facilitate evaluation. More details regarding this process are provided in the Appendix~\ref{sec:datasets}.

\noindent\textbf{Implementation Details.} 
All experiments are conducted on a single Nvidia A100 GPU.
%
For extracting CLIP features, we use the OpenCLIP ViT-B/16 model \jz{add citation}.
For dynamic semantics, we leverage the Qwen2-VL-7B model as the backbone MLLM to generate time-varying captions, and use e5-mistral-7b~\cite{wang2023improving} to encode them into embeddings. Following LangSplat~\cite{qin2024langsplat}, we also train an autoencoder to compress the feature dimension. The CLIP and the text features are compressed into 3 and 6 dimensions, respectively.

\begin{figure*}
    \centering
    \includegraphics[width=1\linewidth]{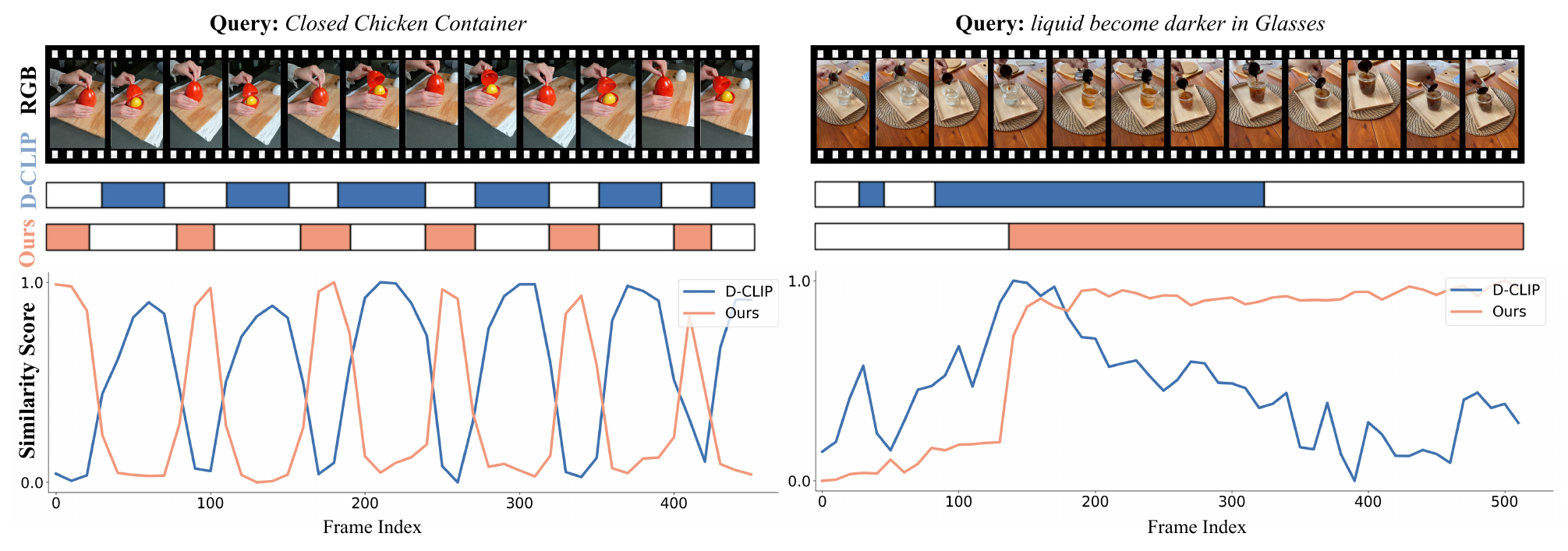}
    \caption{Visualization of time-sensitive querying results between  Deformable CLIP and ours. The bottom row depicts the cosine similarity across frames, rescaled to (0,1) for direct comparison, while the horizontal bars indicate frames identified as relevant time segments. We observed that the CLIP-based method cannot understand dynamic semantics correctly, while our method recognizes them. }
    \label{fig:sim-time-plot}
\end{figure*}
\begin{figure*}
    \centering
    \includegraphics[width=1\linewidth]{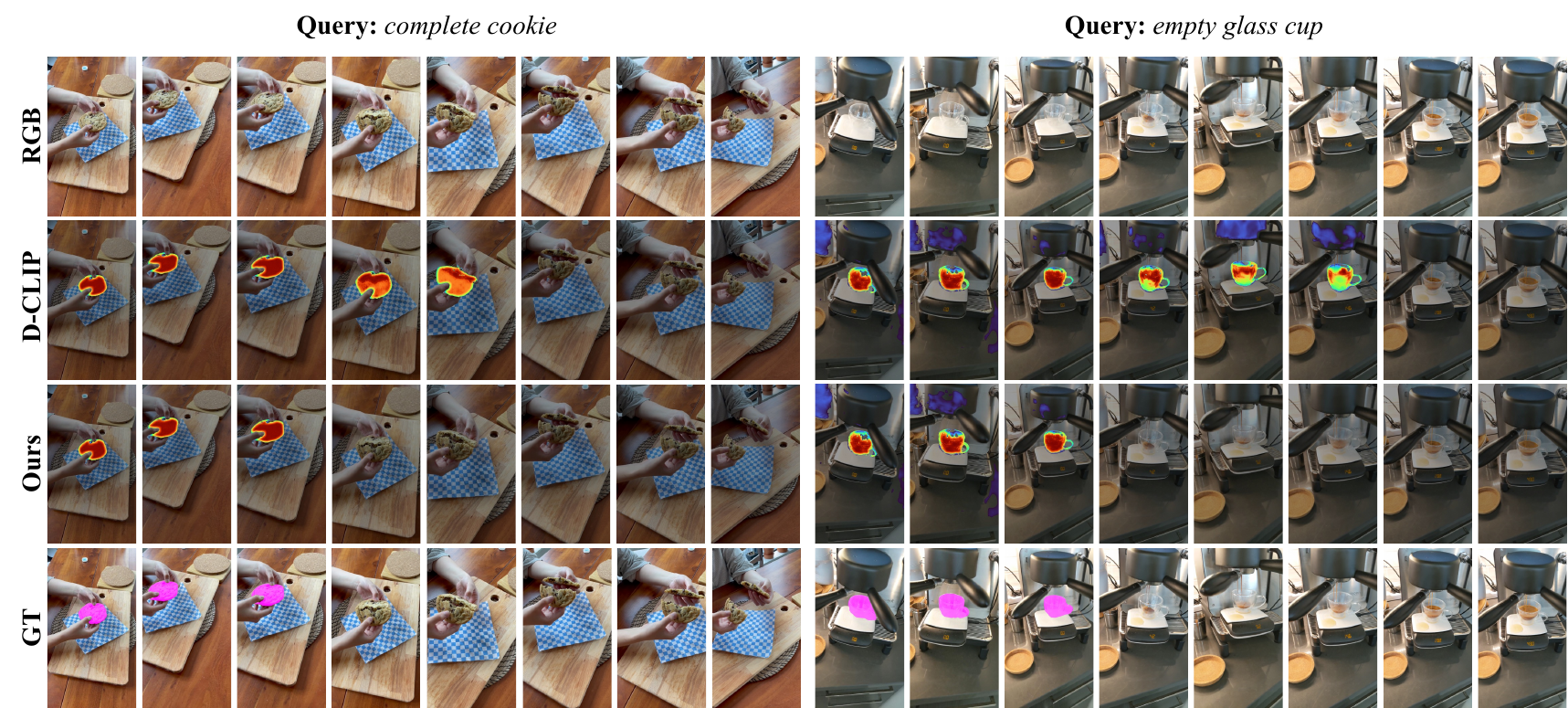}
    \caption{Comparison of time-sensitive query mask. We compare time-sensitive query masks between Deformable CLIP and ours. The CLIP-based method fails to identify time segments accurately, especially at the demarcation points during state transitions. }
    \label{fig:query-visualization}
\end{figure*}

\noindent\textbf{Baselines.} 
Due to the absence of publicly available models for 4D language feature rendering, we use several 3D language feature rendering methods as baselines for evaluating time-agnostic querying, including \Langsplat and \Featuredgs. 
We also incorporate segmentation-based techniques, such as Gaussian Grouping~\cite{ye2025gaussian}, to assess semantic mask generation quality in our approach. Inspired by Segment Any 4D Gaussians~\cite{ji2024segment}, we enhance Gaussian Grouping to adapt to dynamic scenes.


Given the lack of dynamic language field rendering methods, we consider two additional baselines besides LangSplat for time-sensitive querying: Deformable CLIP and Non-Status Field. Deformable CLIP only utilizes the time-agnostic semantic fields of our method, which first trains a 4D-GS model to learn dynamic RGB fields, and then learns static CLIP fields on these pre-trained RGB fields.  The Non-Status Field method utilizes both the time-agnostic semantic field
and the time-sensitive semantic field of our method while removing the status deformable network. Instead, it directly learns a deformation field $\Delta f$.


\noindent\textbf{Metrics.} 
For time-agnostic querying, we evaluate performance using mean accuracy (mAcc) and mean intersection over union (mIoU), calculated across all frames in the test set.
For time-sensitive querying, we evaluate temporal performance using an accuracy metric, defined as $\mathrm{Acc} = {n_{correct}}/{n_{all}}$, where $n_{correct}$ and $n_{all}$ represent the number of correctly predicted frames and the total frames in the test set, respectively. To assess segmentation quality, we adopt the metric from~\cite{zhang2020does} and define $\mathrm{vIoU} = \frac{1}{|S_u|}\sum_{t\in S_i}\mathrm{IoU}(\hat s_t,s_t)$, where $\hat s_t$ and $s_t$ are the predicted and ground truth masks at time $t$, and $S_u$ and $S_i$ are the sets of frames in the union and intersection.



\subsection{Main Results}
\label{subsec:main_results}

\textbf{Time-Agnostic Querying.}
Table \ref{tab:static-main-results} shows our results on two datasets. Our approach achieves the highest mIoU and mAcc scores, demonstrating strong segmentation performance across both datasets. 
In contrast, other methods struggle to capture object movement and shape changes, leading to worse performance on dynamic objects. 


\noindent\textbf{Time-Sensitive Querying.}
We perform dynamic querying on the HyperNeRF dataset, with Acc and vIoU results presented in Table~\ref{tab:dynamic-main-results}.
Our approach outperforms not only the LangSplat method but also the Deformable CLIP and Non-Status Field approaches. Specifically, our method achieves accuracy improvements of 29.03\% and 3.25\% and vIoU gains of 28.04\% and 4.19\%, respectively. 
Our approach introduces a multimodal object-wise video prompting method that surpasses traditional CLIP-based techniques. In comparison to Deformable CLIP, our time-varying semantic field effectively integrates spatial and temporal information. This ensures fluidity and coherence in semantic state transitions, underscoring the importance of MLLM video prompting (Section~\ref{subsec:mllm-video-prompting}). Additionally, when compared to the Non-Status Field method, our approach highlights the significance of status modelling by introducing a status deformable network (Section~\ref{subsec:status-deformable}), which enhances the model’s capability to handle complex, evolving states and further solidifies the robustness and versatility of our method in capturing nuanced dynamics.

\noindent\textbf{Visualization.}
To demonstrate our learned time-sensitive language field, we applied PCA to reduce the dimensionality of the learned semantic features, producing a 3D visualization as shown in Figure~\ref{fig:teaser}. Our method better captures the dynamic semantic features of objects and renders consistent features accurately. In Figure~\ref{fig:sim-time-plot}, we illustrate the change in query-frame similarity scores over time for time-sensitive queries, comparing our approach to a CLIP-based method. As shown, CLIP, which is optimized for static image-text alignment, struggles to capture the most relevant time segments within dynamic video semantics, whereas our method successfully identifies these segments.
In Figure~\ref{fig:query-visualization}, we present specific query masks. We observe that the CLIP-based approach fails to accurately capture time segments, especially at transition points in object states. For example, CLIP cannot reliably detect subtle transitions, such as when a cookie has just cracked or when a glass cup has started dripping coffee. In contrast, our method effectively identifies these nuanced changes, demonstrating its capability to handle dynamic state transitions accurately.
\noindent\subsection{Ablation Studies}

\textbf{Multimodal Prompting.}
We evaluate the quality of generated captions using different combinations of textual and visual prompting methods. To quantify this, we defined a metric, $\Delta_{\mathrm{sim}}=\overline{score}_{\mathrm{pos}} - \overline{score}_{\mathrm{neg}}$, where $\overline{score}_{\mathrm{pos}}$ and $\overline{score}_{\mathrm{neg}}$ represent the average cosine similarity scores between query and caption features, encoded by the e5 model, for positive and negative samples, respectively. A higher $\Delta_{sim}$ indicates a stronger distinction between positive and negative examples, suggesting that the generated caption more effectively captures the spatiotemporal dynamics and semantic features of objects in the scene.
Table~\ref{tab:aba-visual-prompt} shows that utilizing all three visual prompting strategies maximizes the MLLM’s focus on target objects.
As shown in Table~\ref{tab:aba-text-prompt}, incorporating pre-generated video-level motion descriptions
resulting in a 0.87\% improvement. 
Furthermore, adding image prompts enables a more accurate description.


\noindent\textbf{State Numbers.}
Table~\ref{table:ablation-centers} shows the ablation results of the status number $K$.
We observe that an appropriate increase in $K$ led to better results, with 
$K=3$ achieving the optimal performance, which was adopted in our experiments.
\section{Conclusion}

We present 4D LangSplat, a novel approach to constructing a dynamic 4D language field that supports both time-agnostic and time-sensitive open-vocabulary queries within evolving scenes. Our method leverages MLLMs to produce high-quality, object-specific captions that capture temporally consistent semantics across video frames. This enables 4D LangSplat to overcome the limitations of traditional vision feature-based approaches, which struggle to generate precise, object-level features in dynamic contexts. 
By incorporating multimodal object-wise video prompting, we obtain pixel-aligned language embeddings as training supervision.  
Furthermore, we introduce a status deformable network, which enforces smooth, structured transitions across limited object states. Our experimental results across multiple benchmarks demonstrate that 4D LangSplat achieves state-of-the-art performance in dynamic scenarios.


\section*{Acknowledgements} 
The work is supported in part by the National Key R\&D Program of China under Grant 2024YFB4708200 and National Natural Science Foundation of China under Grant U24B20173, and in part by US NIH grant R01HD104969.

{
    \small
    \bibliographystyle{ieeenat_fullname}
    \bibliography{main}
}

\appendix
\setcounter{page}{1}
\maketitlesupplementary

\section{Datasets}
\label{sec:datasets}
Since there are no publicly available ground truth segmentation mask labels for the \Hypernerf and \Neud datasets, nor annotations tailored for time-sensitive querying, we adopt the annotation pipeline outlined in Segment Any 4D Gaussians~\cite{ji2024segment} and manually annotate the mask labels ourselves. Specifically, we leverage the Roboflow platform alongside the SAM (Segment Anything Model) framework for interactive annotation.



For the HyperNeRF dataset, where data is captured with a monocular camera, we select one frame every four frames as the training set. From the remaining data, we annotate a subset as the test set to ensure no overlap between the two sets.
For the Neu3D dataset with 21 camera views, one is reserved for testing, and the remaining 20 are used for training, aligning with the 4D-GS~\cite{wu20244d} setting. To evaluate on the Neu3D dataset, we annotate every 20 frames from the test views.

\section{Implementation Details}
\textbf{Multimodal Object-Wise Video Prompting.}
For Multimodal Object-Wise Video Prompting, we utilize the largest SAM-defined semantic levels as mask inputs for the Multimodal Large Language Models (MLLMs). The prompting process is outlined in Table~\ref{tab:prompts-detail}, which provides the specific prompts used for MLLM prompting. 
For visual prompting, we employ a red contour line with a radius of 2 to delineate object boundaries. Additionally, we apply Gaussian blur with a radius of 10 and convert the images to grayscale mode to achieve gray-level augmentation. These techniques enhance the effectiveness of the visual input during the prompting process.

\begin{table}[h!]
\centering
\begin{tabular}{p{0.1\textwidth} p{0.3\textwidth}}
\toprule
Video prompts  & Image prompts  \\
\midrule
 I highlighted the objects I want you to describe in red outline and blurred the objects that don't need you to describe. First please determine the object highlighted in red line in the video. Then briefly summarize the transformation process of this object.  & You have an understanding of the overall transformation process of the object: \{video prompt\}. Now, I have provided you with images extracted from this process. Please describe the specific state of the object(s) in the given image, without referring to the entire video process.  Avoid describing states that you can't infer directly from the picture. Avoid repeating descriptions in context. For example, if the context suggests the object is moving up and down but the image shows it is just moving down, explicitly only state that the object is in a moving down state. If the context suggests the object is breaking but the image shows it is complete right now, explicitly only state that the object appears to be complete. If context tells you something changes from green to blue, but it's blue in this image, just state that the object is blue.\\
\bottomrule
\end{tabular}
\vspace{-5pt}
\caption{Details of Text prompts}
\vspace{-5pt}
\label{tab:prompts-detail}
\end{table}

\noindent\textbf{Autoencoder.} Following LangSplat~\cite{qin2024langsplat}, we employ two autoencoders to compress the high-dimensional CLIP feature (512-dimension) and LLM feature (4096-dimension) separately.
Specifically, two MLPs are used to compress 512-dimensional CLIP features and 4096-dimensional video features to 3 and 6 dimensions, respectively. The autoencoders are optimized with L2 loss. To enhance stability, a cosine similarity loss is also included as a regularization. 

\begin{table}[t]
\renewcommand\tabcolsep{12pt}
\centering
\begin{tabular}{lc}
\toprule
Method &  FPS  \\
\midrule
\Gaussiangrouping & 1.47 \\
Ours-agnostic & \textbf{5.24} \\
Ours-sensitive & 4.05 \\
\bottomrule
\end{tabular}
\caption{Query Performance Comparison.}
\vspace{-10pt}
\label{tab:query-efficiency}
\end{table}

\begin{table*}[t]
\renewcommand\tabcolsep{8pt}
\centering
\begin{tabular}{ccccccc}
\toprule
\multirow{2}{*}{Method} & \multicolumn{2}{c}{americano} & \multicolumn{2}{c}{chickchicken} & \multicolumn{2}{c}{split-cookie} \\
\cmidrule(lr){2-3} \cmidrule(lr){4-5} \cmidrule(lr){6-7}
& mIoU(\%) & mAcc(\%) & mIoU(\%) & mAcc(\%) & mIoU(\%) & mAcc(\%) \\
\midrule
\Featuredgs & 34.65 & 62.96 & 47.21 &	87.22 & 47.03 &	68.25 \\
\Gaussiangrouping & 61.77 & 71.31 & 34.65	& 75.52 & 72.71 &	96.56 \\
\Langsplat & 72.08	& 97.61 & 75.98	& 97.86 & 76.54	& 97.32 \\
\midrule
Ours & \textbf{83.48} 	& \textbf{98.77} & \textbf{86.50} 	& \textbf{98.81}  & \textbf{90.04} 	& \textbf{98.67}  \\
\midrule
\multirow{2}{*}{Method} & \multicolumn{2}{c}{espresso} & \multicolumn{2}{c}{keyboard} & \multicolumn{2}{c}{torchocolate} \\
\cmidrule(lr){2-3} \cmidrule(lr){4-5} \cmidrule(lr){6-7}
& mIoU(\%) & mAcc(\%) & mIoU(\%) & mAcc(\%) & mIoU(\%) & mAcc(\%) \\
\midrule
\Featuredgs & 24.04 & 80.13 & 42.14 &	80.98 & 24.71 &	64.58 \\
\Gaussiangrouping & 32.45 & 82.46 & 42.44 & 74.15 & 58.95 &	85.52 \\
\Langsplat & 82.93	& \textbf{98.66} & 72.42 &	\textbf{96.75} & 69.55 & 98.09 \\
\midrule
Ours & \textbf{83.52} & 97.95 & \textbf{79.53}	& 95.71 & \textbf{71.79}	& \textbf{98.10}  \\
\bottomrule
\end{tabular}
\vspace{-5pt}
\caption{Comparison of mean IoU and mean Accuracy for various methods on the \Hypernerf datasets.}
\vspace{-5pt}
\label{tab:static-main-results-hypernerf}
\end{table*}
\begin{table*}[t]
\renewcommand\tabcolsep{8pt}
\centering
\begin{tabular}{ccccccc}
\toprule
\multirow{2}{*}{Method} & \multicolumn{2}{c}{coffee martini} & \multicolumn{2}{c}{cook spinach} & \multicolumn{2}{c}{cut roasted beef} \\
\cmidrule(lr){2-3} \cmidrule(lr){4-5} \cmidrule(lr){6-7}
& mIoU(\%) & mAcc(\%) & mIoU(\%) & mAcc(\%) & mIoU(\%) & mAcc(\%) \\
\midrule
\Featuredgs & 30.23 & 84.74 & 41.50 & 95.59 & 31.66 &	91.07  \\
\Gaussiangrouping & 71.37 & 97.34 & 46.45 & 93.79 & 54.70	& 93.25 \\
\Langsplat & 67.97	& 98.47 & 78.29 & 98.60 & 36.53 & 97.04 \\
\midrule
Ours & \textbf{85.16} &	\textbf{99.23} & \textbf{85.09} &	\textbf{99.38} & \textbf{85.32}	& \textbf{99.28} \\
\midrule
\multirow{2}{*}{Method} & \multicolumn{2}{c}{flame salmon} & \multicolumn{2}{c}{flame steak} & \multicolumn{2}{c}{sear steak} \\
\cmidrule(lr){2-3} \cmidrule(lr){4-5} \cmidrule(lr){6-7}
& mIoU(\%) & mAcc(\%) & mIoU(\%) & mAcc(\%) & mIoU(\%) & mAcc(\%) \\
\midrule
\Featuredgs & 54.33 & 77.13 & 27.27 &	88.23 & 24.78 & 85.94 \\
\Gaussiangrouping & 35.72 & \textbf{94.69} & 36.92 &	95.96 & 54.44	& 95.27 \\
\Langsplat & 66.01	& 82.16 & 64.05 & 97.77 & \textbf{78.29} & 98.60 \\
\midrule
Ours &\textbf{ 89.88} &	94.35 & \textbf{88.44}	& \textbf{98.27} & 76.78	& \textbf{99.38} \\
\bottomrule
\end{tabular}
\vspace{-5pt}
\caption{Comparison of mean IoU and mean Accuracy for various methods on the \Neud dataset.}
\vspace{-10pt}
\label{tab:static-main-results-neu3d}
\end{table*}

\noindent\textbf{Training Details.}
Our training pipeline is structured into four stages, progressively refining the model for robust performance in dynamic 4D language field construction.
1) In the initial stage, we train a static Gaussian field to reconstruct the RGB channel of static scenes. This provides a foundation for modeling the visual appearance of the scene. 2) Next, we incorporate semantic information into the static Gaussian field without introducing deformable networks. Semantic features are embedded into the scene by minimizing an $L_1$ loss, ensuring accurate representations of the static scene's semantics. 3) In the third stage, we extend the model to dynamic RGB scenes by introducing non-semantic deformation fields. Leveraging the approach of 4D-GS~\cite{wu20244d}, we employ deformable networks to learn temporal and motion-based deformations that capture spatial and temporal dynamics for RGB scenes. 4) For time-agnostic semantic rendering, we refine the semantic features from the second stage while keeping the deformable network parameters fixed.
For time-sensitive semantic rendering, we jointly train the status deformable network and the state prototype features to refine and model dynamic semantics effectively.
For all datasets, the iterations for four stages are 3000, 1000, 10000, and 10000. The learning rates for the deformable network and the state prototype features are set to \(1.6 \times 10^{-4}\) and \(2.5 \times 10^{-3}\), respectively.
Other training parameters remain consistent with those used in 4D-GS.

\section{More Quantitative Results}
In Table~\ref{tab:static-main-results-hypernerf} and Table~\ref{tab:static-main-results-neu3d}, we present a detailed evaluation of time-agnostic querying performance on the HyperNeRF and Neu3D datasets, respectively. Our method achieves a mean IoU exceeding 85\% across all scenarios,  outperforming the baseline methods in most scenes for both mean IoU and mean accuracy. These results underscore the robustness of our approach, demonstrating its ability to deliver superior segmentation accuracy and reliability compared to existing methods, even in dynamic scenes.

Table~\ref{tab:query-efficiency} further compares the runtime efficiency of our method with the baseline on the HyperNeRF dataset. The comparison encompasses the total time required for rendering semantic features and conducting open-vocabulary queries. Our method demonstrates significant advantages over the Gaussian Grouping approach, achieving faster runtime for both time-agnostic and time-sensitive queries. These findings validate our method as an efficient and scalable solution for handling open-vocabulary queries in dynamic 4D scenes.

\begin{figure*}[t]
    \centering
    \includegraphics[width=1\linewidth]{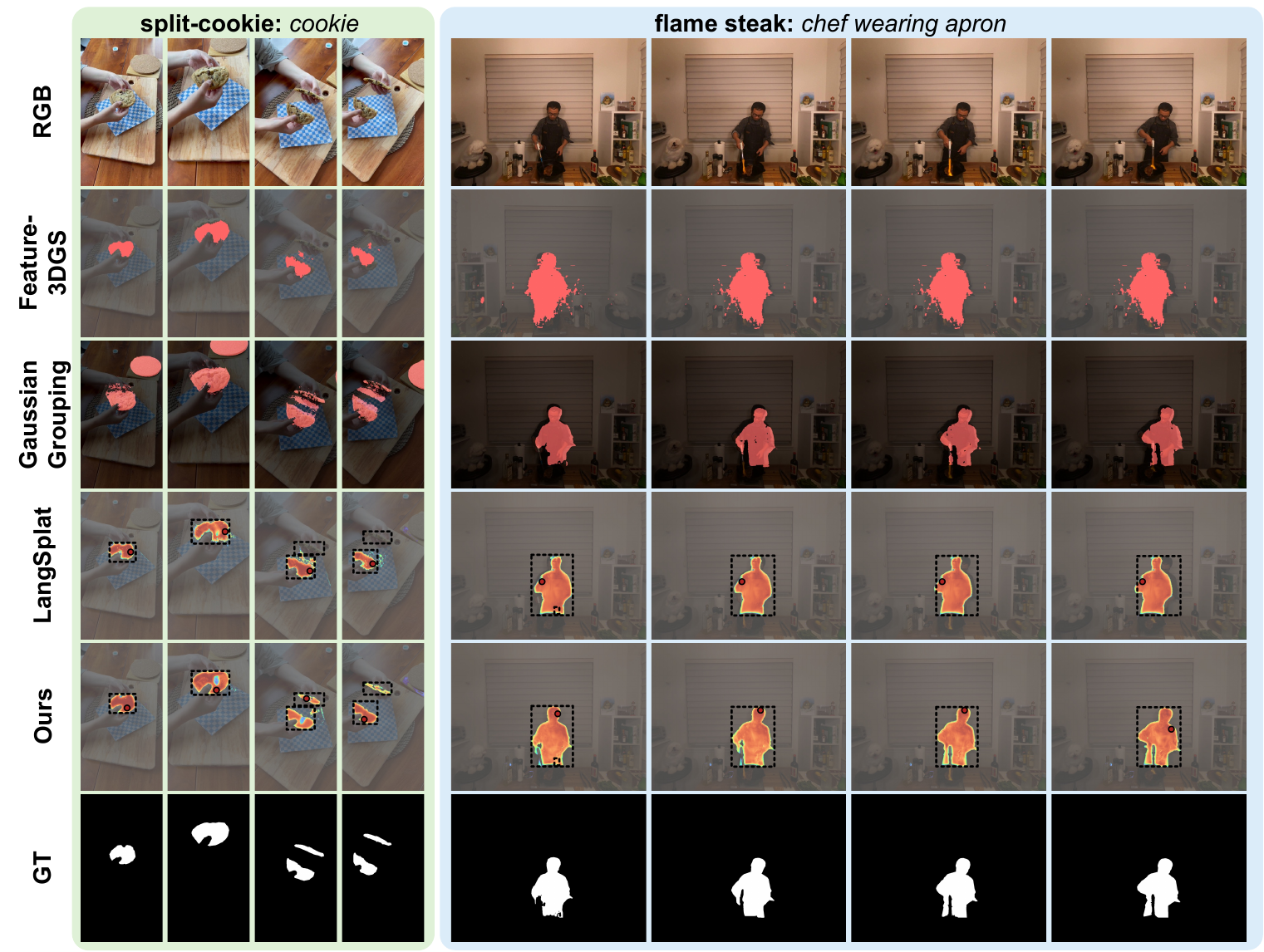}
    \caption{Visualization of time-agnostic querying results on \Hypernerf and \Neud datasets. }
    \label{fig:more-query-visualization}
\end{figure*}

\section{More Visualization Results}
Figure~\ref{fig:more-query-visualization} illustrates visualization results for time-agnostic querying. As depicted, our method demonstrates superior accuracy in capturing objects that correspond to semantic descriptions, compared to other methods. Furthermore, it effectively tracks the spatial dynamics of these objects across different temporal steps, showcasing its effectiveness in handling dynamic scenarios.

\section{MLLM-based Embeddings} 

Since our method utilizes MLLMs to generate captions, the feature representation capability of the obtained embeddings is inherently limited by the capacity of the MLLMs, which constitutes a limitation of our approach. To verify that our MLLM-based embeddings indeed encode spatial-temporal information, we directly apply the MLLM-based embeddings, without any fine-tuning, to video classification and spatial-temporal action localization tasks using 2D videos. As shown in Tables \ref{tab:video-classification} and \ref{tab:spatial-temporal-localization}, our results demonstrate that, even in a zero-shot setting, the MLLM-based embeddings achieve competitive performance compared to state-of-the-art (SOTA) methods specifically designed for these tasks. This indicates that MLLM-based embeddings inherently capture some spatial-temporal information. However, we also acknowledge that the performance of our approach is ultimately constrained by the representational capacity of the MLLMs.

\begin{table}[t]
\renewcommand\tabcolsep{3pt}
\centering
\begin{tabular}{l|c c c }
\hline
 Method & HMDB51~\cite{kuehne2011hmdb} & UCF101~\cite{soomro2012ucf101} & Kinetics400~\cite{kay2017kinetics} \\
\hline
MLLM & 58.34 & 78.97 & 55.14 \\
IMP~\cite{akbari2023alternating} & 59.1  &  91.5  & 	77.0 \\  
\hline
\end{tabular}
\caption{Accuracy Results (\%) on the Video Classification task.}\label{tab:video-classification}
\end{table}

\begin{table}[t]
\centering
\begin{tabular}{l|c c c}
\hline
Method & VmAP@0.1 & VmAP@0.2 & VmAP@0.5 \\
\hline
MLLM  & 78.13 & 75.78 & 64.38 \\
HIT~\cite{faure2023holistic} & 86.1 & 88.8  & 74.3 \\      
\hline
\end{tabular}
\caption{Spatial-Temporal Action Localization Results (\%) on UCF101~\cite{soomro2012ucf101}.} \label{tab:spatial-temporal-localization}
\end{table}

\end{document}